\documentclass{article}
\usepackage{spconf,amsmath,graphicx,amsfonts}
\usepackage{cite}
\usepackage{booktabs} 
\usepackage{tabularx}
\usepackage{enumitem}
\usepackage{textcomp}
\usepackage{cite}
\usepackage[english]{babel}
\usepackage{mathtools}

\usepackage{ifpdf}
\usepackage{algorithmic}
\usepackage{multirow}
\usepackage{xcolor}
\usepackage{array}
\usepackage{url}
\usepackage[colorlinks=true, urlcolor=blue, linkcolor=red]{hyperref}
\usepackage{comment}
\usepackage{xurl}

\newcommand*{\Scale}[2][4]{\scalebox{#1}{$#2$}}%

\title{Explaining Representation Learning with Perceptual Components}
\name{Yavuz Yarici  \enskip Kiran Kokilepersaud  \enskip Mohit Prabhushankar \enskip Ghassan AlRegib \enskip }

\address { OLIVES at the Center for Signal and Information Processing CSIP,\\ 
School of Electrical and Computer Engineering, Georgia Institute of Technology, Atlanta, GA, USA \\
\{yavuzyarici, mohit.p, alregib\}@gatech.edu   }

\begin{document}
%\ninept
%

\twocolumn[{%

{ \large
\begin{itemize}[leftmargin=2.5cm, align=parleft, labelsep=2cm, itemsep=4ex,]

\item[\textbf{Citation}]{Y. Yarici, K. Kokilepersaud, M. Prabhushankar, G. AlRegib, "Explaining Representation Learning with Perceptual Components," in \textit{2024 IEEE International Conference on Image Processing (ICIP), Abu Dhabi, United Arab Emirates (UAE), 2024.}}

\item[\textbf{Review}]{Date of Acceptance: June 6th 2024}

\item[\textbf{Codes}]{\url{https://github.com/olivesgatech/Explaining-Representation-Learning-with-Perceptual-Components.git}}

\item[\textbf{Bib}]  {@inproceedings\{yarici2024xaiperceptual,\\
    title=\{Explaining Representation Learning with Perceptual Components\},\\
    author=\{Yarici, Yavuz and Kokilepersaud, Kiran, and Prabhushankar, Mohit and AlRegib Ghassan\},\\
    booktitle=\{ IEEE International Conference on Image Processing (ICIP)\},\\
    year=\{2024\}\}}

\item[\textbf{Copyright}]{\textcopyright 2024 IEEE. Personal use of this material is permitted. Permission from IEEE must be obtained for all other uses, in any current or future media, including reprinting/republishing this material for advertising or promotional purposes, creating new collective works, for resale or redistribution to servers or lists, or reuse of any copyrighted component of this work in other works.}

\item[\textbf{Contact}]{

\{yavuzyarici, mohit.p, alregib\}@gatech.edu \\\url{https://ghassanalregib.info/}\\}
\end{itemize}

}}]

\maketitle
\begin{abstract}
Self-supervised models create representation spaces that lack clear semantic meaning.  This interpretability problem of representations makes traditional explainability methods ineffective in this context. In this paper, we introduce a novel method to analyze representation spaces using three key perceptual components: color, shape, and texture. We employ selective masking of these components to observe changes in representations, resulting in distinct importance maps for each. In scenarios, where labels are absent, these importance maps provide more intuitive explanations as they are integral to the human visual system. Our approach enhances the interpretability of the representation space, offering explanations that resonate with human visual perception. We analyze how different training objectives create distinct representation spaces using perceptual components. Additionally, we examine the representation of images across diverse image domains, providing insights into the role of these components in different contexts.

\end{abstract}

\section{Introduction}

\label{sec:intro}

In recent years, machine learning models have significantly advanced the state-of-the-art for image classification tasks. These models have complex designs and contain billions of parameters,  which complicates their interpretability, effectively turning them into 'black box' machines. This lack of transparency is particularly problematic in critical sectors like healthcare and autonomous technologies, where trust in decision-making processes is paramount \cite{extractingcausal}. The need for trustworthy machine learning models sparked interest in the field of explainable artificial intelligence (XAI) that focuses on understanding the decisions of AI systems \cite{Linardatos2020,alregib2022explanatory}. 

Currently, many strategies exist to understand the decisions of supervised models. In the domain of supervised learning, the model functions by mapping meaningful features to meaningful labels denoted as $f: X \rightarrow y$. In the supervised setting, explanations try to understand the decision of the model. These explanations are comprehensible to humans primarily because the labels themselves carry significance for us.

However, in many application domains, obtaining high-quality labels is often a scarce and costly endeavor \cite{focal}. This situation has driven the adoption of self-supervised algorithms. The effectiveness of these algorithms stems from their ability to optimize based on representations instead of explicit annotations, significantly reducing the costs associated with data labeling.

\begin{figure}[t]
\centering
\includegraphics[scale=0.38]{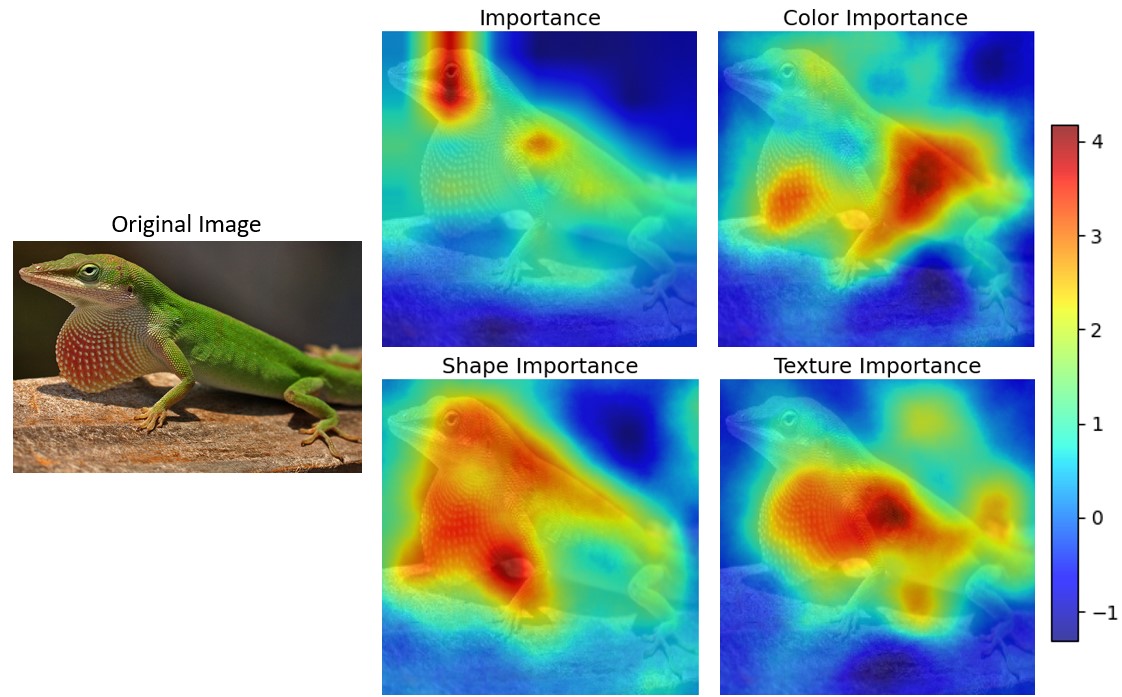}

\caption{This figure shows the overall importance score and importance scores for Color, Shape, and Texture for SimCLR ImageNet pre-trained encoder.  ResNet50 is used as a backbone for the encoder. The image is taken from ImageNet.  }

\label{fig:lizard}
\end{figure}

Despite these advantages, challenges arise when evaluating self-supervised learning frameworks. Common assessment methods, which focus on performance in downstream tasks from feature extractors, offer only limited insights into the features models use. Such evaluations miss critical differences, as similar accuracies in tasks do not guarantee that models base their representations on identical features \cite{selfsupervisedanalysis}. This highlights the crucial role of explainability in representation learning, emphasizing the need for a deeper comprehension of how these models process and interpret data.

However, there is a noticeable gap in research on explaining representations.  This is primarily because self-supervised models function differently from supervised models. Instead of mapping inputs to labels, they map inputs to a representation space, denoted as $f: X \rightarrow \mathbb{S}$.  Self-supervised models often generate a representation space that lacks clear semantic meaning in a way that is incomprehensible to humans.  This interpretability problem of representations makes traditional explainability methods ineffective in this context. Addressing this gap in research is vital for advancing our understanding of self-supervised models and enhancing their transparency and applicability in real-world scenarios.

To navigate this, recent developments in XAI are geared towards unsupervised settings. Recent works \cite{Wickstrm2023,crabbé2022labelfree,corpus}, have focused on identifying features that, upon removal, significantly shift the transformed representation away from its original representation. This approach aims to determine the input feature importance, bringing a degree of interpretability and understanding to the representation space. Particularly, \cite{Wickstrm2023}  focuses on explaining the representation space in unsupervised models by randomly masking the input image and observing changes in the representation. The motivation for masking is that when parts of the image are removed the representation space shifts in a manner that reflects the importance of that region. This gives us a pixel-level importance map for our representation. 

Although these explanations try to bring interpretability to the field, importance maps from these models are not straightforward, due to the lack of semantic context to guide which features should be deemed important. For instance, Fig. \ref{fig:lizard} displays an importance map for a representation space produced by a SimCLR-trained encoder. This map primarily highlights the lizard's head and certain parts of its back. However, it is unclear why these specific areas are emphasized over others.

To address this, we devise a novel approach to analyze representation spaces based on three fundamental perceptual components: color, shape, and texture. We achieve this by selectively masking each component and observing the resulting changes in their representations. This gives us a separate importance map for each component. Fig. \ref{fig:lizard} presents our importance maps for color, shape, and texture using the same lizard image. Our primary motivation for employing perceptual components in explaining representation space is that these components offer a more intuitive explanation in the absence of labels since they are integral to the human visual system. For example, as seen in Fig. \ref{fig:lizard}, the lizard's head is highlighted in both the shape importance and overall importance maps but not in the others. Thus, we can infer that the reason why the head is highlighted in the overall importance map is mostly because of its shape. These explanations can make representation-based explanations more meaningful and comprehensible to humans.

We then assess the relative importance of these components in models trained for various objectives.  Our findings reveal considerable variations in the importance scores assigned to different perceptual components, depending on the training objectives. This highlights the way different training approaches result in distinct representation spaces, where each perceptual component is weighted differently. Additionally, we analyze the importance scores of these components across various datasets. This analysis shows that the significance of specific perceptual components varies across different image domains. The contributions of this paper are as follows:
\begin{enumerate}
    \item We introduce a method to create a pixel-level importance map for perceptual components, particularly color, shape, and texture.
    \item We analyze the representation space with perceptual components in models with diverse training objectives where we undercover significant differences based on training objectives.
    \item We analyze the representations of images from different image domains based on color, shape, and texture.    
\end{enumerate}

\begin{figure*}[t]
\centering
\includegraphics[scale=0.55]{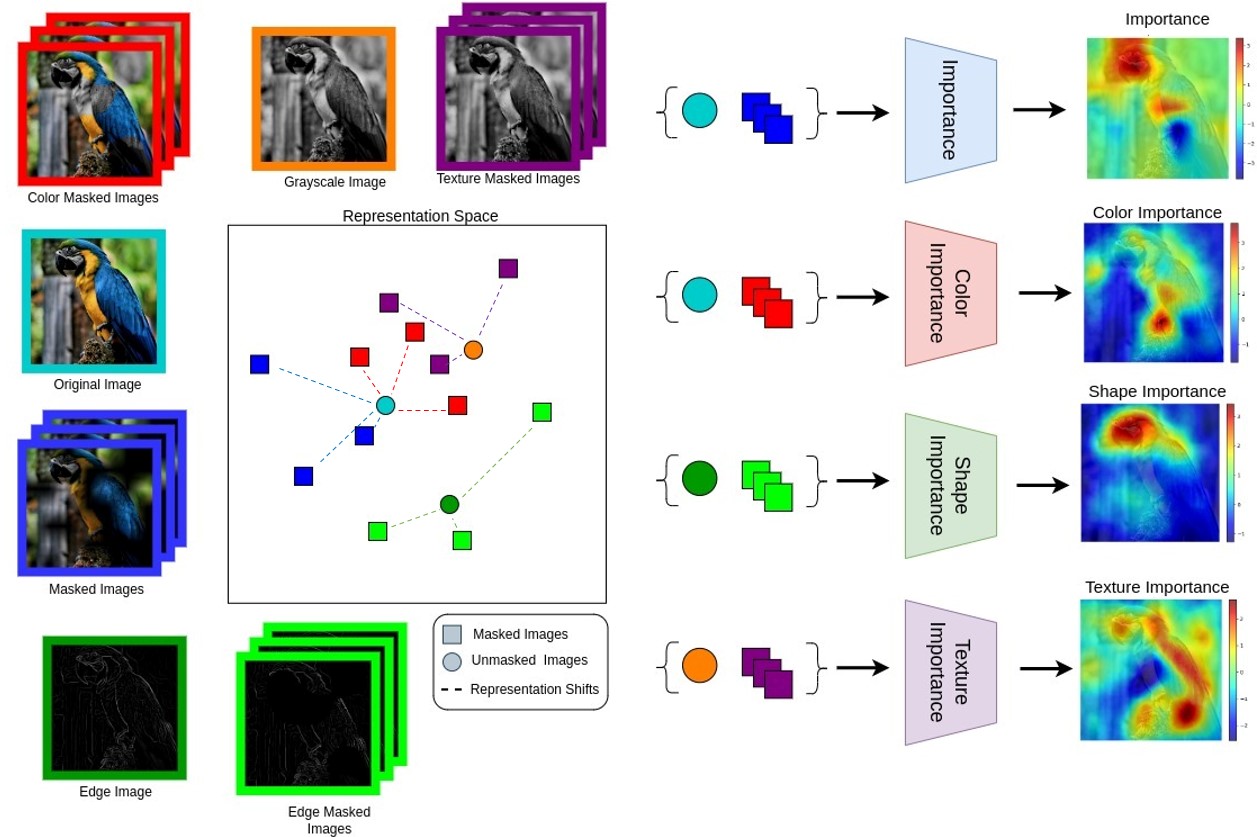}

\caption{ This diagram illustrates how the importance map for Color, Shape, and Texture is produced. Circles represent the unmasked images and rectangles represent the masked images. Cosine similarities of unmasked and masked images are used for importance map generation for each component.}

\label{fig: method}
\end{figure*}

\section{Related Works}

\textbf{Occlusion-Based Explainability:} Supervised learning algorithms use occlusion-based methods to produce explanations of model decisions. RISE\cite{rise} proposed to explain the model decision by masking random parts of the input image and observing the model's predictions to determine the important parts of the input. \cite{meanngfulpert} optimizes a spatial perturbation mask to significantly impact the model's output, while \cite{extremeperturb} extends this concept by identifying perturbations that have the most substantial effect on the network's output within a defined area.

In contrast, the field of unsupervised learning has seen relatively fewer developments in explainability.  RELAX\cite{Wickstrm2023}  focuses on explaining the representation space in unsupervised models by randomly masking the input image and observing changes in the representation. \cite{crabbé2022labelfree} developed a method to determine label-free feature importance and label-free example importance, which identifies key features and training examples used by models to build representations during inference. Another innovative approach is COCOA \cite{corpus}, which explains unsupervised models based on similarity within a representation space. This method allows users to select specific samples (corpus and foil) and inquire about the features that make the representation of their chosen sample similar to the corpus but different from the foil. Although these methods provide an overall explanation,  they differ from our approach as they do not offer any explanation based on perceptual components such as color, shape, and texture.

\textbf{Perceptual Components for Machine Learning}: Over the years, many studies have been done to investigate the effect of perceptual components like color, shape, and texture on machine learning models. \cite{Susstrunk} analyzed how CNNs process color information and the influence of color-sensitive features on the network.  Similarly, \cite{colorinfo} explored the effects of color variations caused by different illumination and sensor characteristics on CNN classification performance. Adding to this, \cite{colorrobust} investigated the influence of color distortions on deep neural network performance in image classification. \cite{colorimportance} empirically investigates the importance of colors in object recognition models.  \cite{geirhos2022imagenettrained} demonstrated a bias in ImageNet-trained CNNs towards recognizing textures over shapes. \cite{hvs} took a different approach by constructing a humanoid visual engine that separately processes shape, texture, and color features from images. While these studies collectively enhance our understanding of the influence of shape, color, and texture on CNNs, they are distinct from our work in that they do not offer a pixel-level explanation of how these perceptual components affect the representation space in CNNs.

\section{Methodology}

\begin{figure*}[t]
\centering
\includegraphics[scale=0.50]{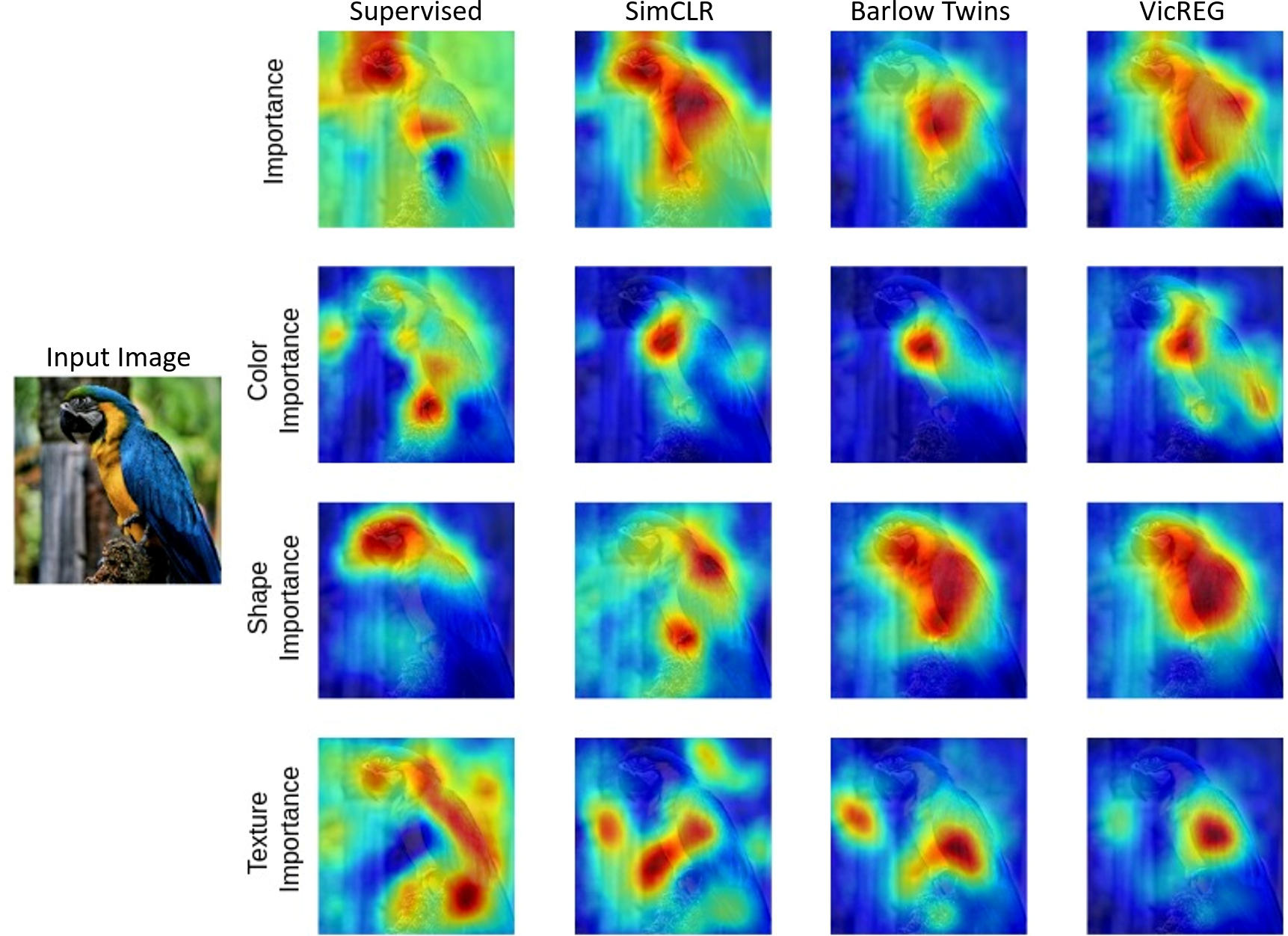}    

\caption{This figure shows importance scores and importance scores for Color, Shape, and Texture for Supervised, SimCLR, Barlow Twins, and VicReg models. Red indicates high values and blue indicates low values.  }
 
\label{fig: importances}
\end{figure*}

\vspace{-1mm}
To create importance maps for a representation space, we use a masking strategy similar to \cite{Wickstrm2023}. Our approach identifies the most informative parts of an image by observing how changes in the image affect its feature representation. The underlying principle is that masking out informative parts of the input image should lead to a significant shift in the representation space. Our methodology to create importance maps for each perceptual component follows the same motivation. When informative parts of the perceptual components are masked out, the representation should significantly shift in representation space which gives us information about how important that perceptual component is. Our methodology is illustrated in Fig. \ref{fig: method}. 

\subsection{Importance Map Creation}

Let $h=f(\mathbf{X}) \in \mathbb{R}^D$ is an encoder that maps our input image $\mathbf{X} \in \mathbb{R}^{H\times W}$ to the representation space. For masking, we apply random binary mask $\textbf{M} \in [0,1]^{H \times W}$ with distribution $D$. Masking operation can be written as $ \mathbf{X} \odot \textbf{M}$ where  $\odot$ is an element-wise multiplication

$\Bar{h} = f(\mathbf{X} \odot \textbf{M})$ represent our masked representation. If the masking operation removes informative pixels for our representation space, the similarity between $h$ and $\Bar{h}$ should be small. Based on this we define the importance of pixel $(i,j)$ as:
\vspace{-1mm}
\begin{equation}
    R_{ij}=E[s(h,\Bar{h})M_{ij}]
\end{equation}

\vspace{-1mm}
For computational ease, we approximate the expectation in the equation by the sample mean as : 
\vspace{-1mm}
\begin{equation}
    R_{ij}=\frac{1}{N} \sum_{n=1}^{N} s(h,\Bar{h_n}) M_{ij}(n)
\end{equation}

\vspace{-1mm}
We use cosine similarity as our metric to measure representation shifts. To generate a random mask, we first sample N binary masks of a smaller size than the original image, setting each element to 1 with a probability $p$. We then upscale all masks by a factor of 256 using bicubic interpolation. After that, we apply a random resize crop to make it the original image size. We sample 8000 masks and calculate the importance map for each pixel with equation 2.

\subsection{Importance Map for Perceptual Components}
\subsubsection{Color:}

To create the color importance map, we mask the original image with the grayscale image our masking operation can be denoted as: 
\begin{equation}
    \mathbf{X_{MC}}= (\mathbf{X} \odot \textbf{M}) + (\mathbf{X_{grayscale}} \odot (1-\textbf{M}))
\end{equation}

where $\mathbf{X_{grayscale}}$ is a grayscale transformed input image and $\mathbf{X_{MC}}$ is color masked image. To calculate color importance, we use equation 2 applying the cosine similarity between the representations of the original input and the color-masked image:

\vspace{-1mm}
\begin{equation}
      R_{ij}^{Color}=\frac{1}{N} \sum_{n=1}^{N} s( f(\mathbf{X}), f(\mathbf{X_{MC}})) M_{ij}(n)  
\end{equation}
\vspace{-1mm}
\subsubsection{Shape:}

To create the shape importance map we first apply edge detection to extract shape information for the input image. We use the Canny Edge detection method \cite{Canny1986}. Then, we mask edge images with binary masks.  The masking operation can be denoted as:  
\begin{equation}
\mathbf{X_{MS}}= \mathbf{X_{EdgeImage}} \odot \mathbf{M}  
\end{equation}

where $\mathbf{X_{EdgeImage}}$ is the output of edge detection and $\mathbf{X_{MS}}$ is edge masked image. To calculate shape importance, we use equation 2 applying the cosine similarity between the edge image and the edge-masked image:

\begin{equation}
\Scale[0.9]{R_{ij}^{Shape}=\frac{1}{N} \sum_{n=1}^{N} s( f(\mathbf{X_{EdgeImage}}), f(\mathbf{X_{MS}})) M_{ij}(n)}   
\end{equation}

\subsubsection{Texture:}

To generate the texture importance map, we initially transform the input image into a grayscale. Next, we apply a mask to the grayscale image using Gaussian blur. The masking operation can be expressed as:

\begin{equation}
\mathbf{X_{MT}}=  (\mathbf{X_{grayscale}} \odot \textbf{M}_{t} + (\mathbf{X_{blur}} \odot (1-\textbf{M}_t))    
\end{equation}

where $\textbf{X}_{grayscale}$ is the grayscale transformation to the input image, $\mathbf{X_{blur}}$ is gaussian blurred grayscale image  and $\textbf{X}_{MT}$ is texture masked image. Here we use the mask with the addition of edge image and normal mask $\textbf{M}_t=\textbf{ M}  \lor \mathbf{X_{EdgeImage}}$ for masking where $\lor$ is logic element wise OR operator between two binary inputs. This masking ensures that the edges are not affected by blur operation. 

To understand the process here, we need to first define the texture. Texture is defined as a function of the ordered spatial variation in pixel intensities (gray values)\cite{texture}. Hence by applying Gaussian blur, we remove the texture information from the masked patches.  

To calculate the final texture importance,  we use equation 2, applying the cosine similarity between the grayscale image and the grayscale-masked image:

\begin{equation}
\Scale[0.9]{     R_{ij}^{Texture}=\frac{1}{N} \sum_{n=1}^{N} s( f(\mathbf{X_{grayscale}}), f(\mathbf{X_{MT}})) M_{ij}(n) }  
\end{equation}

\section{Experiments}

We conduct numerous experiments to analyze the representation space with perceptual components. We report both quantitative and qualitative results. Our evaluation encompasses a range of pre-trained models. Each model is pre-trained with different training objectives. We use ImageNet pre-trained models from the VISSL library \cite{goyal2021vissl}. All pre-trained models employ the same ResNet-50 architecture. We use four different models to analyze the representation space created by different models: Supervised, SimCLR\cite{simclr}, Barlow Twins\cite{barlow}, and VICReg\cite{vicreg}. These models are chosen due to their differences in their training objectives which results in distinct representation spaces.

Furthermore, we experiment with diverse datasets to explore how different image domains exhibit distinct characteristics in their representation spaces. For each experiment, we randomly sample 1,000 images from the dataset. We make use of five datasets in our analysis: ImageNet\cite{imagnet}, Cars\cite{cars}, Birds\cite{birds}, Flowers\cite{flower}, and Cure-OR\cite{cureor}. We choose ImageNet to represent a broad spectrum of general image domains. To examine the differences across specific image domains, we choose Cars, Birds, and Flowers data set. From Cure-OR, we select images featuring a single object without any background or noise, aiming to study representations of images that contain only one object.

\subsection{Qualitative Evaluation }
Overall importance scores and importance scores for each perceptual component are displayed in  Fig. \ref{fig: importances}. The supervised model primarily focuses on the head. In contrast, SimCLR and VicReg highlight both the body and head. Barlow Twins mainly highlights the body. This variation illustrates that the importance attributed to different parts of an image significantly depends on the model's training method.

In terms of color importance, the Supervised model highlights the bird's entire body, whereas other models predominantly focus locally on the lower body.  

For shape importance, the Supervised model gave a much more targeted importance map by highlighting only the head. VicReg and Barlow twins on the other hand indicated that the shape of the whole body is important. 
The regions highlighted for overall importance in VicReg and Barlow Twins models align closely with those for shape importance.

Regarding texture importance, all models highlight the bird's blue feathers, with VicReg doing so more precisely.

\subsection{Quantitative Evaluation }

 As part of our numerical analysis, we examined the correlation between the overall importance map and the importance maps of individual perceptual components. This correlation score serves as a measure of agreement between the overall importance map and the maps of individual components. We refer to these scores as 'agreement scores', and it is calculated by this formula for each component:

\begin{equation}
Z_{component}={\sum_{i=0}^{N} \sum_{j=0}^{N} R_{ij}^{component} \cdot  R_{ij}^{Importance}}    
\end{equation}

\begin{table}[ht]
\centering
\begin{tabular}{@{}cccc@{}}
\toprule
\multicolumn{4}{c}{Agreement Scores for Different Models on ImageNet}   
\\ \midrule
\multicolumn{1}{c}{Model} & \multicolumn{1}{c}{Color} & \multicolumn{1}{c}{Shape} & \multicolumn{1}{c}{Texture} \\
\midrule
Supervised ResNet50 & {0.414}          & 0.350        & 0.261    \\
SimCLR  ResNet50     & 0.309              & {0.557}              & 0.239        \\
VICReg  ResNet50     & \textbf{0.489}              & {0.770}              & 0.477        \\
Barlow Twins  ResNet50     & 0.470              & \textbf{0.803}              & \textbf{0.490}        \\

\bottomrule
\end{tabular}
\vspace*{.2mm}
\caption{ Analysis of agreement scores for different models on samples from ImageNet dataset. }
\label{tab:aggrement_models}
\end{table}

\begin{table}[ht]
\centering
\begin{tabular}{@{}cccc@{}}
\toprule
\multicolumn{4}{c}{Aggrement Scores  for Different Datasets}   
\\ \midrule
\multicolumn{1}{c}{Model} & \multicolumn{1}{c}{Color} & \multicolumn{1}{c}{Shape} & \multicolumn{1}{c}{Texture} \\
\midrule

ImageNet\cite{imagnet}      & 0.309              & {0.557}              & 0.239        \\
Birds\cite{birds} & 0.300          & 0.588        & 0.319    \\
Cars\cite{cars} & 0.330          & 0.589        & 0.165    \\
Flowers\cite{flower} & 0.355          & 0.550        & 0.252    \\
CURE-OR\cite{cureor} & 0.420 &	0.742 &	0.488 \\

\bottomrule
\end{tabular}
\vspace*{.1mm}
\caption{Analysis of agreement scores for SimCLR on different datasets. }
\label{tab:agreement_data}
\end{table}

For each perceptual component, the generated importance maps reveal the relative significance of each pixel for that component. if pixels that are highly important in the overall map also exhibit high importance in a specific perceptual component importance map, that specific component is important for our representation. Hence having a high agreement score with one perceptual component indicates that that component is important for the representation of the sample.

We first analyze the agreement scores for different models. The results for agreement scores for different models are shown in Table \ref{tab:aggrement_models}. Notably, the agreement scores for color and texture in the SimCLR pre-trained model are lower compared to those in the supervised-trained model. This can be attributed to SimCLR's training process, which incorporates augmentations such as color jitter and Gaussian blur, making it more invariant in color and texture. Consequently, we observe reduced agreement for these attributes. In contrast, the SimCLR model exhibits significantly higher agreement scores for shape compared to the supervised model. This indicates that a decrease in agreement with certain features, such as color and texture, may correspond to an increase in agreement with others, like shape. On the other hand, the agreement scores for VICReg and Barlow Twins are considerably higher than those for SimCLR and Supervised models. This can be ascribed to the training processes of Barlow Twins and VICReg, where they utilize objective functions to mitigate feature collapse resulting from augmentations. Overall, the agreement scores for different perceptual components vary significantly based on the training objectives. This variation indicates that different training objectives create distinct representation spaces, assigning varying degrees of importance to each perceptual component.
We also examined the agreement scores across various datasets for the SimCLR pre-trained model. These results are detailed in Table \ref{tab:agreement_data}. Agreement scores for the CURE-OR dataset are notably higher, which aligns with expectations since CURE-OR typically features one small object per image, leading to a higher overlap among importance maps. A comparison of texture agreement scores between datasets reveals a distinct pattern: the birds dataset exhibits a significantly higher texture agreement compared to the cars dataset, with ImageNet and Flowers positioned in the middle. This indicates that textured structures are more crucial for identifying birds. This outcome aligns with expectations, as birds typically display more textured features compared to smoother objects like cars and flowers. Furthermore, the agreement score for color in the flower dataset surpasses that of the cars, birds, and ImageNet datasets. This emphasizes the significance of color in representing flowers, indicating that the color component plays a crucial role in distinguishing flowers from other objects.

\section{Conclusion}

In this paper, we introduce a novel method for analyzing representation spaces in machine-learning models using three key perceptual components: color, shape, and texture. By employing selective masking techniques, we generate distinct importance maps for each component, enhancing the interpretability of representation learning in alignment with human visual perception. We analyze the influence of training methodologies and domain characteristics on representation spaces. We demonstrate the variations in the importance assigned to different perceptual components based on training objectives, illustrating how training approaches result in unique representation spaces with varied emphasis on each perceptual component. We also demonstrated that the importance of specific perceptual components varies across different image domains, demonstrating their significant role in model explainability.

% \vfill
% \pagebreak

% \vfill\pagebreak

\bibliographystyle{IEEEbib}
\bibliography{ref}

\end{document}